\documentclass[sigconf]{acmart}
\AtBeginDocument{%
  \providecommand\BibTeX{{%
    \normalfont B\kern-0.5em{\scshape i\kern-0.25em b}\kern-0.8em\TeX}}}

\copyrightyear{2021} 
\acmYear{2021} 
\setcopyright{acmlicensed}\acmConference[CIKM '21]{Proceedings of the 30th ACM International Conference on Information and Knowledge Management}{November 1--5, 2021}{Virtual Event, QLD, Australia}
\acmBooktitle{Proceedings of the 30th ACM International Conference on Information and Knowledge Management (CIKM '21), November 1--5, 2021, Virtual Event, QLD, Australia}
\acmPrice{15.00}
\acmDOI{10.1145/3459637.3482139}
\acmISBN{978-1-4503-8446-9/21/11}

\usepackage{bm}
\usepackage{graphicx}
\usepackage{subfigure}
\usepackage{balance}
\begin{document}
\fancyhead{}
\title{MDFEND: Multi-domain Fake News Detection}

\author{Qiong Nan\textsuperscript{1,2,3}, Juan Cao\textsuperscript{1,2,*}, Yongchun Zhu\textsuperscript{1,2}, Yanyan Wang\textsuperscript{1,2}, Jintao Li\textsuperscript{1}}
\affiliation{%
\institution{\textsuperscript{1}Key Lab of Intelligent Information Processing of Chinese Academy of Sciences (CAS), Institute of Computing Technology, CAS, Beijing, China 
\textsuperscript{2}University of Chinese Academy of Sciences, Beijing, China \\
\textsuperscript{3}State Key Laboratory of Media Convergence Production Technology and Systems, Beijing, China}
\streetaddress{}
\city{}
\state{}
\country{}
\postcode{}
\thanks{*Juan Cao is the corresponding author.}
}
\email{{nanqiong19z, caojuan, zhuyongchun18s, wangyanyan18s, jtli}@ict.ac.cn}

\renewcommand{\shortauthors}{Qiong Nan, et al.}

\begin{abstract}
  Fake news spread widely on social media in various domains, which lead to real-world threats in many aspects like politics, disasters, and finance. Most existing approaches focus on single-domain fake news detection (SFND), which leads to unsatisfying performance when these methods are applied to multi-domain fake news detection. As an emerging field, multi-domain fake news detection (MFND) is increasingly attracting attention. However, data distributions, such as word frequency and propagation patterns, vary from domain to domain, namely domain shift. Facing the challenge of serious domain shift, existing fake news detection techniques perform poorly for multi-domain scenarios. Therefore, it is demanding to design a specialized model for MFND. In this paper, we first design a benchmark of fake news dataset for MFND with domain label annotated, namely Weibo21, which consists of 4,488 fake news and 4,640 real news from 9 different domains. We further propose an effective \textbf{\underline{M}}ulti-\textbf{\underline{d}}omain \textbf{\underline{F}}ak\textbf{\underline{e}} \textbf{\underline{N}}ews \textbf{\underline{D}}etection Model (MDFEND) by utilizing a domain gate to aggregate multiple representations extracted by a mixture of experts. The experiments show that MDFEND can significantly improve the performance of multi-domain fake news detection. Our dataset and code are available at \url{https://github.com/kennqiang/MDFEND-Weibo21}.
\end{abstract}

\ccsdesc[500]{Computing methodologies~Transfer learning}
\ccsdesc[300]{Information systems~Data mining}

\keywords{fake news detection; multi-domain; social media; dataset}


\maketitle

\section{Introduction}
In recent years, with the rapid popularization of the Internet, social media such as Sina Weibo~\cite{SinaWeibo} and Twitter~\cite{Twitter}, has become an important source to acquire news. However, it also serves as an ideal platform for fake news dissemination. According to Weibo's 2020 annual report on refuting rumors~\cite{deepfake}, there are 76,107 fake news pieces treated by authority all year round.
Since fake news can cause devastating consequences to individuals and society, fake news detection is a critical problem that needs to be addressed.

To solve the problem, a variety of approaches have been proposed, and most of them~\cite{castillo2011information,kwon2013prominent,ma2016detecting,ma2017content,jin2017president} focus on single-domain fake news detection (SFND), e.g., politics, health. However, for a certain domain, the amount of fake news can be extremely limited. Therefore, based on such inadequate single-domain data, the performance of these detection models is unsatisfying. In practical scenarios, the real-world news platforms release various news in different domains everyday~\cite{silva2021embracing}. Therefore, it is promising to solve the data sparsity problem and improve the performance of all domains by exploiting data from multiple domains, called multi-domain fake news detection (MFND).

However, the data distributions vary from domain to domain, called domain shift~\cite{pan2009survey,zhuang2020comprehensive}.
First, different domains have different word usage, for example, the most commonly used words in military news are ``navy", ``army", etc, while in educational news, ``students", ``university", ``teacher", etc; Second, the propagation patterns vary a lot in different domains~\cite{silva2021embracing}.
Facing the problem of serious domain shift, MFND can therefore be quite challenging.
Moreover, some domains only contain very little labeled data, and this phenomenon further increases the difficulty in MFND, which remains unsolved yet with existing methods. 

To study MFND, we build a comprehensive dataset Weibo21, which contains news from 9 domains, including Science, Military, Education, Disasters, Politics, Health, Finance, Entertainment and Society. Every domain contains news content, released timestamp, corresponding pictures and comments. Since fake news is intentionally created for financial or political gain, it often contains opinionated and inflammatory language. It is reasonable to exploit linguistic features of news content to detect fake news~\cite{shu2017fake}.
Timestamp and comments are also included with news, as timestamp can be used to do sequential analyses~\cite{ma2016detecting}. And comments can provide auxiliary signals especially when the posts don't contain abundant information~\cite{zhang2021mining}. In the end, Weibo21 contains 4,488 fake news and 4,640 real news from 9 different domains.

Due to the lack of systematic work of MFND, we adopt several multi-domain learning baselines~\cite{wang2018eann,ma2018modeling,qin2020multitask,silva2021embracing} and evaluate these multi-domain methods as well as several popular single-domain fake news detection methods~\cite{kim5882convolutionalneuralnetworksforsentence,ma2016detecting} on our proposed dataset Weibo21. In addition, we propose a simple but effective Multi-domain Fake News Detection Model, namely MDFEND, which utilizes a domain gate to aggregate multiple representations extracted by a mixture of experts. Our experiments demonstrate the significant effectiveness improvement of the proposed MDFEND compared with the aforementioned baselines.

The main contributions of this work are summarized into three folds: (1) We construct Weibo21, an MFND dataset. To the best of our knowledge, this data repository is the first MFND dataset collected from one platform and contains the richest domains. (2) We proposed a simple but effective method named MDFEND for MFND. (3) We systematically evaluate MFND performance of different methods on our proposed Weibo21 dataset.
\vspace{-0.2cm}
\section{Related Work}
\textbf{I. Fake News Detection}. Many approaches have been proposed to tackle the challenges of fake news detection. Earlier studies used hand-craft features~\cite{castillo2011information,kwon2013prominent,giachanou2019leveraging,ajao2019sentiment}. 
Some recent research works use propagation patterns for structural modeling~\cite{ma2016detecting,ma2017detect,guo2018rumor,song2019ced}, others jointly used both textual and visual features for multi-modal modeling~\cite{jin2017multimodal,qi2019exploiting}.
Ma et al.~\cite{ma2018detect} and Li et al.~\cite{li2019rumor} incorporated related tasks to assist fake news detection.
Wang et al.~\cite{wang2018eann} adopted the idea of the minimax game to extract event-invariant (domain-invariant) features, but neglected domain-specific features. Silver et al.~\cite{silva2021embracing} proposed to jointly preserve domain-specific as well as cross-domain knowledge to detect fake news from different domains, but they didn't make full use of the domain information explicitly. 
\textbf{II. Multi-domain (multi-task) learning}. The thought of multi-domain (multi-task) learning is to jointly learn a group of domains (tasks), which has been proved effective in many applications~\cite{ma2018modeling,qin2020multitask,zhao2019multiple,zhu2019aligning,zhu2021learning}. 
These researches focus on capturing relationships of different tasks with multiple representations. And each task is reinforced by the interconnections, including inter-task relevance difference.
However, these multi-domain (multi-task) frameworks aren't fit for fake news detection. Therefore, it is necessary to design an appropriate and effective method for MFND.

\textbf{Datasets}. A few datasets have been constructed for fake news detection, including LIAR~\cite{wang2017liar}, CoAID~\cite{cui2020coaid}, FakeHealth~\cite{dai2020ginger}, Twitter16~\cite{ma2016detecting} and Weibo~\cite{ma2016detecting, zhang2021mining}, however, they don't have multi-domain information. FakeNewsNet~\cite{shu2020fakenewsnet} only contains two domains, including Politifact and GossipCop, which are insufficient for multi-domain fake news detection.
Therefore, an appropriate multi-domain fake news dataset is desperately in need.
\vspace{-0.35cm}
    

\begin{table}[htbp]
  \centering
  \small
  \setlength{\abovecaptionskip}{-0.010cm}
  \caption{Data Statistics of Weibo21}
    
    \resizebox{7.6cm}{1.514cm}{
    \begin{tabular}{lccccc}
    \toprule
    \textbf{domain} & \textbf{Science} & \textbf{Military} & \textbf{Education} & \textbf{Disasters} & \textbf{Politics} \\
    \midrule
    \textbf{real} & 143   & 121   & 243   & 185   & 306 \\
    \textbf{fake} & 93    & 222   & 248   & 591   & 546 \\
    \midrule
    \textbf{all} & 236   & 343   & 491   & 776   & 852 \\
    \midrule
    \midrule
    \textbf{domain} & \textbf{Health} & \textbf{Finance} & \textbf{Entertainment} & \textbf{Society} & \textbf{All} \\
    \midrule
    \textbf{real} & 485   & 959   & 1000  & 1198  & 4640 \\
    \textbf{fake} & 515   & 362   & 440   & 1471  & 4488 \\
    \midrule
    \textbf{all} & 1000  & 1321  & 1440  & 2669  & 9128 \\
    \bottomrule
    \end{tabular}%
    }
  \label{tab:data_statistic}%
\end{table}%
\vspace{-0.6cm}
\section{Weibo21: A New Dataset for MFND}

In this section, we describe the process of data collection for Weibo21, a multi-domain fake news dataset in Chinese, as well as how we determine the domain category of news pieces. Further, we conduct a preliminary data analysis based on news content, the most straightforward clue in fake news detection, to show some domain differences explicitly.
\vspace{-0.4 cm}
\subsection{Data Collection}
We collect both fake and real news from Sina Weibo~\cite{SinaWeibo} ranging from \textbf{December 2014} to \textbf{March 2021}.

\textbf{Initial Data Crawling.} For fake data, we collect news pieces judged as misinformation officially by Weibo Community Management Center~\cite{WeiboService}. For real data, we gather real news pieces in the same period as the fake news, which have been verified by NewsVerify~\cite{Newsverify} (a platform that focuses on discovering and verifying suspicious news pieces on Weibo).
For each piece of news, we collect (1) the most straightforward information, \textbf{news content}, (2) different modality, {\it i.e.}, \textbf{pictures}, (3) sequential signals, {\it i.e.},\textbf{timestamp}, and (4) social context, {\it i.e.}, \textbf{comments}. 
Further, we gather \textbf{judgement information} for fake news, which can provide evidence to people and increase the credibility of our dataset. 

\textbf{Deduplication.} The raw data contians lots of duplication, which may cause data leakage in training procedure, so we perform deduplication in one-pass clustering.  
Finally, 4,488 pieces of fake news and 4,640 pieces of real news are obtained.
\vspace{-0.4cm}
\subsection{Domain Annotation}
After data collection, we perform categorization on crowd-sourcing. First, in order to work out the categorization criterion, we refer to domain lists from several well-known fact-checking websites, including Zhuoyaoji~\cite{zhuoyaoji}, Liuyanbaike~\cite{liuyanbaike}, Jiaozhen~\cite{jiaozhen}, and Ruijianshiyao~\cite{Ruijianshiyao}, as well as some research papers and reports, including Vosoughi et al.~\cite{vosoughi2018spread}, 2017 Tencent Rumor Governance Report~\cite{Tencentreport} and China Joint Internet Rumor-Busting Platform~\cite{Rumorbusting}. Considering the congruence as well as the appropriateness of granularity, we end with nine domains: \textbf{Science}, \textbf{Military}, \textbf{Education}, \textbf{Disasters}, \textbf{Politics}, \textbf{Health}, \textbf{Finance},  \textbf{Entertainment}, and \textbf{Society}. 
Then, we annotate all pieces of news into the above nine domains. To ensure the fairness of our annotation, 10 experts are employed to manually annotate the news.
At first, the 10 experts annotate all the news independently, i.e. each piece of news is labeled by 10 experts; then they check the annotation with each other; The final domain label can be determined if more than 8 experts choose the same one, otherwise, they will discuss with each other till reaching to an agreement. 
The statistics of the collected datasets are shown in Table \ref{tab:data_statistic}.
\vspace{-0.4 cm}
\subsection{Preliminary Data Analysis}
\begin{figure}
    \setlength{\belowcaptionskip}{-0.5 cm}
    \setlength{\abovecaptionskip}{0 cm}
    \centering
    \subfigure[health]{\includegraphics[scale = 0.0105]{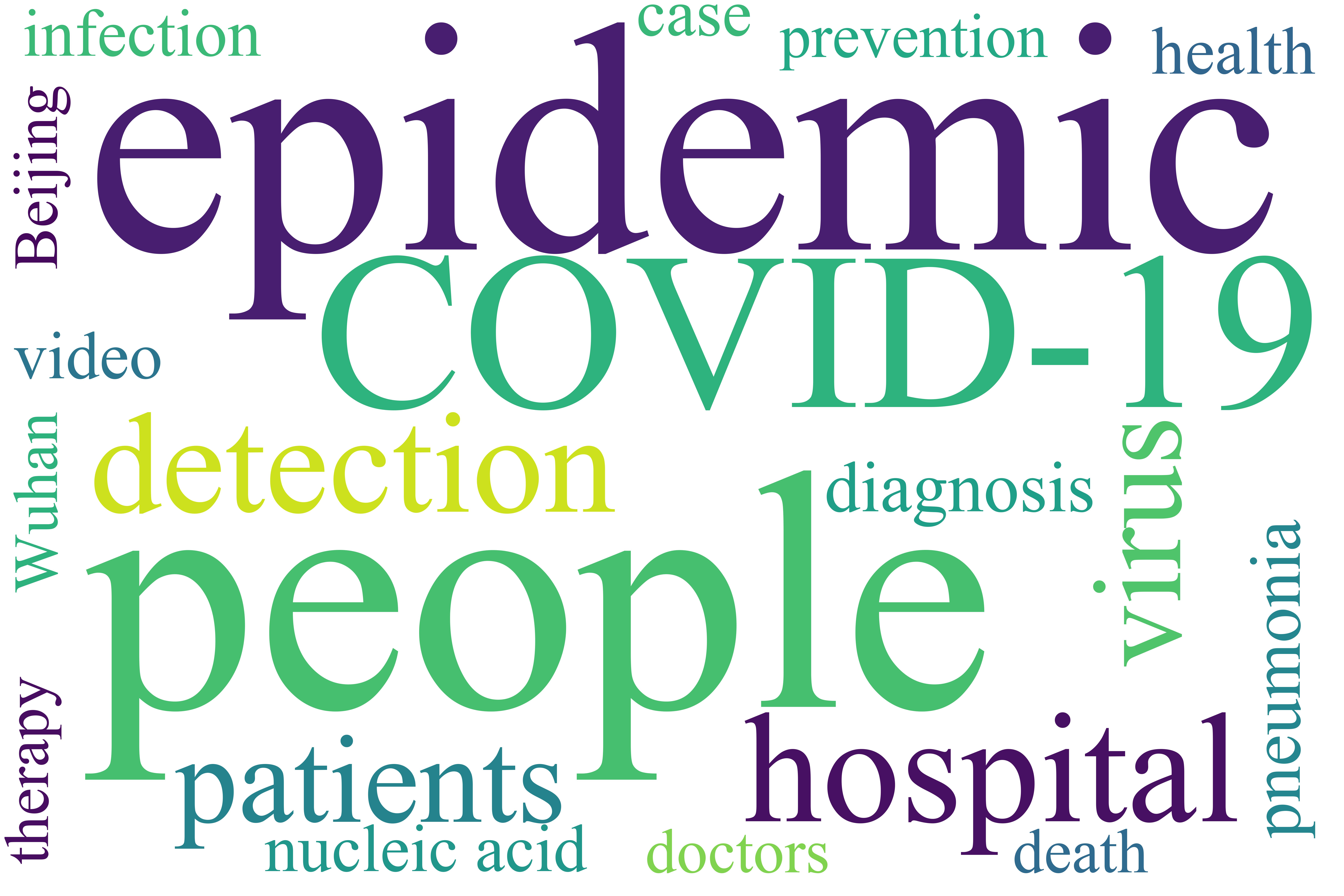}}
    \subfigure[military]{\includegraphics[scale = 0.0105]{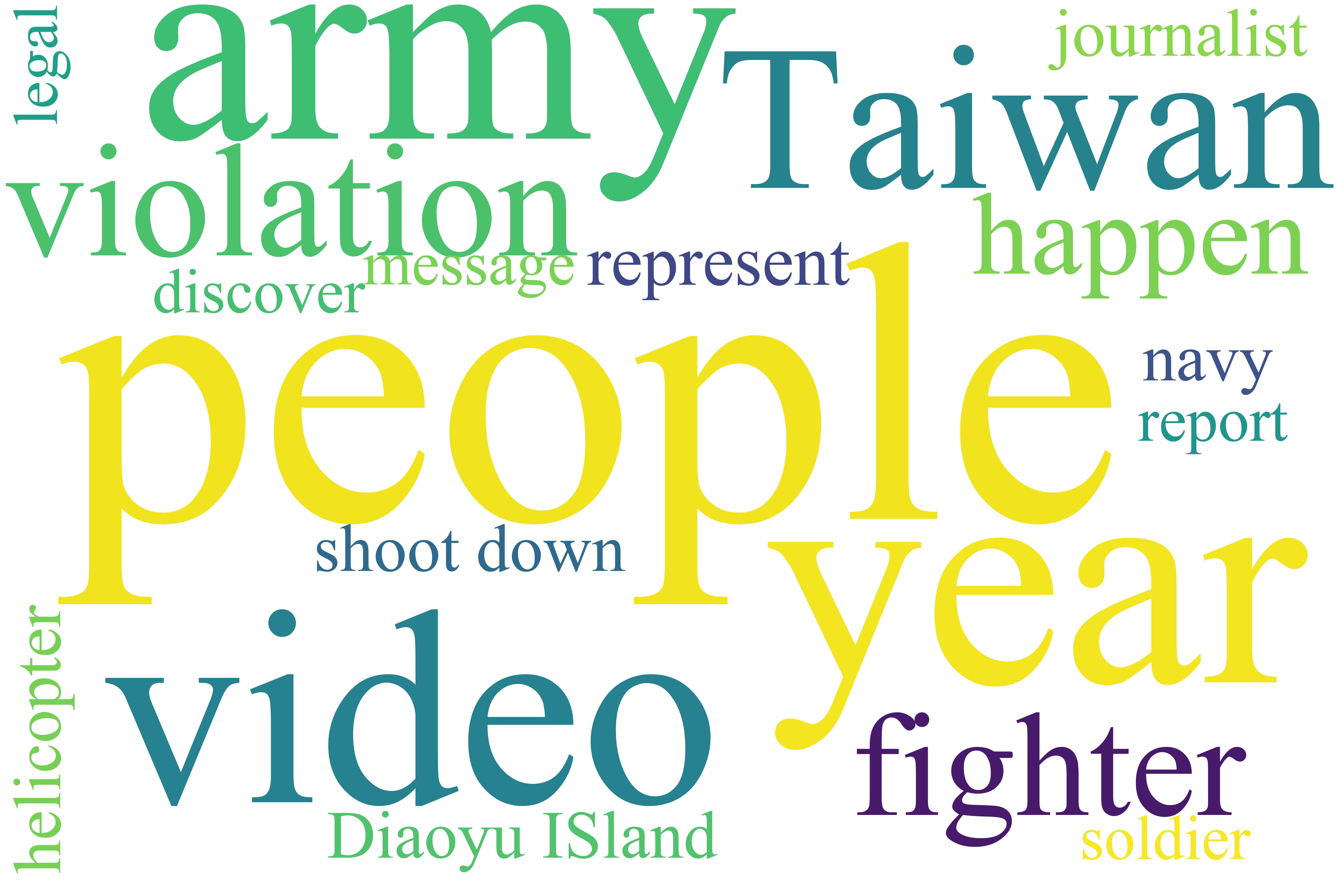}}
    \\
    \centering
    \subfigure[education]{\includegraphics[scale = 0.0105]{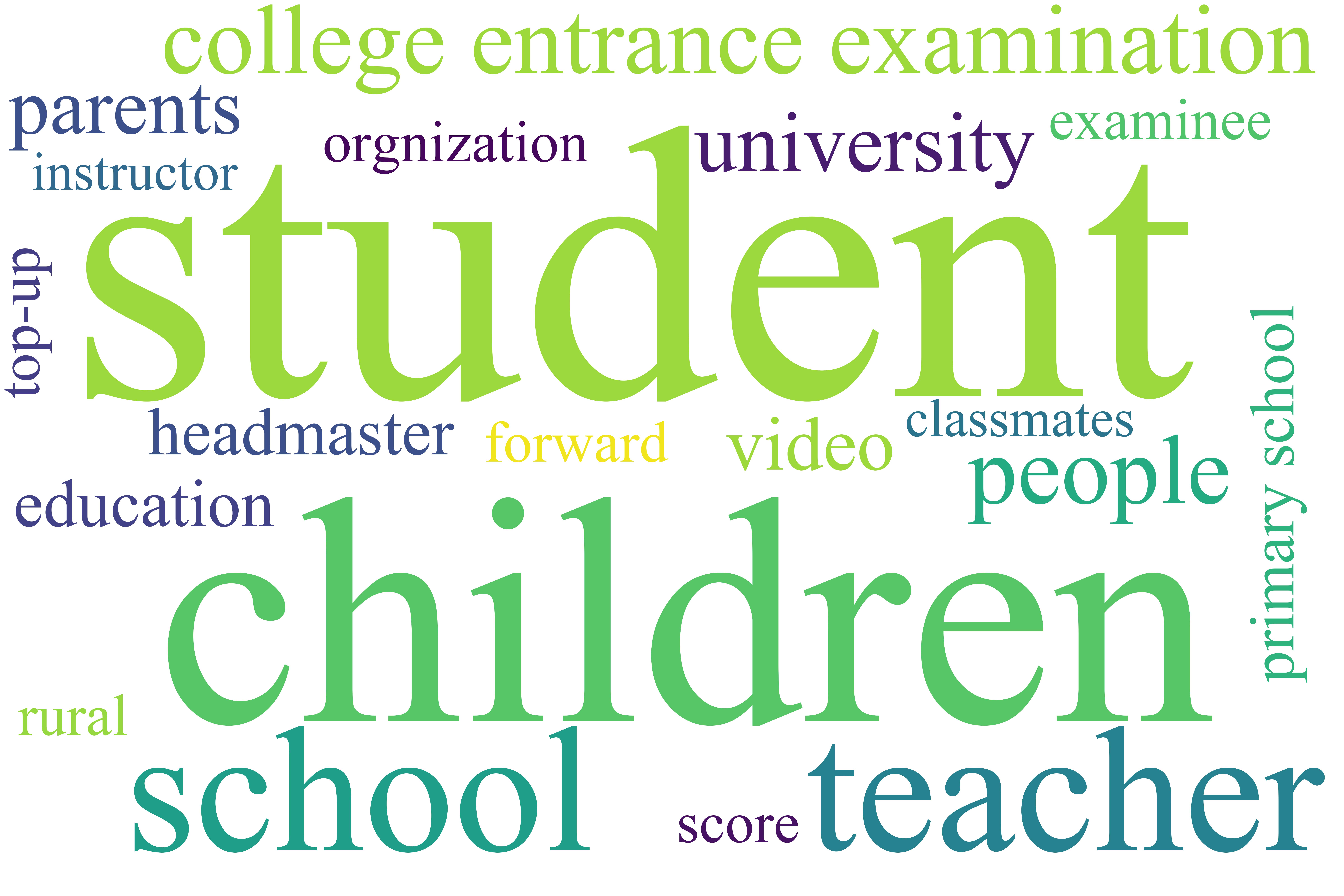}}
    \subfigure[science]{\includegraphics[scale = 0.0105]{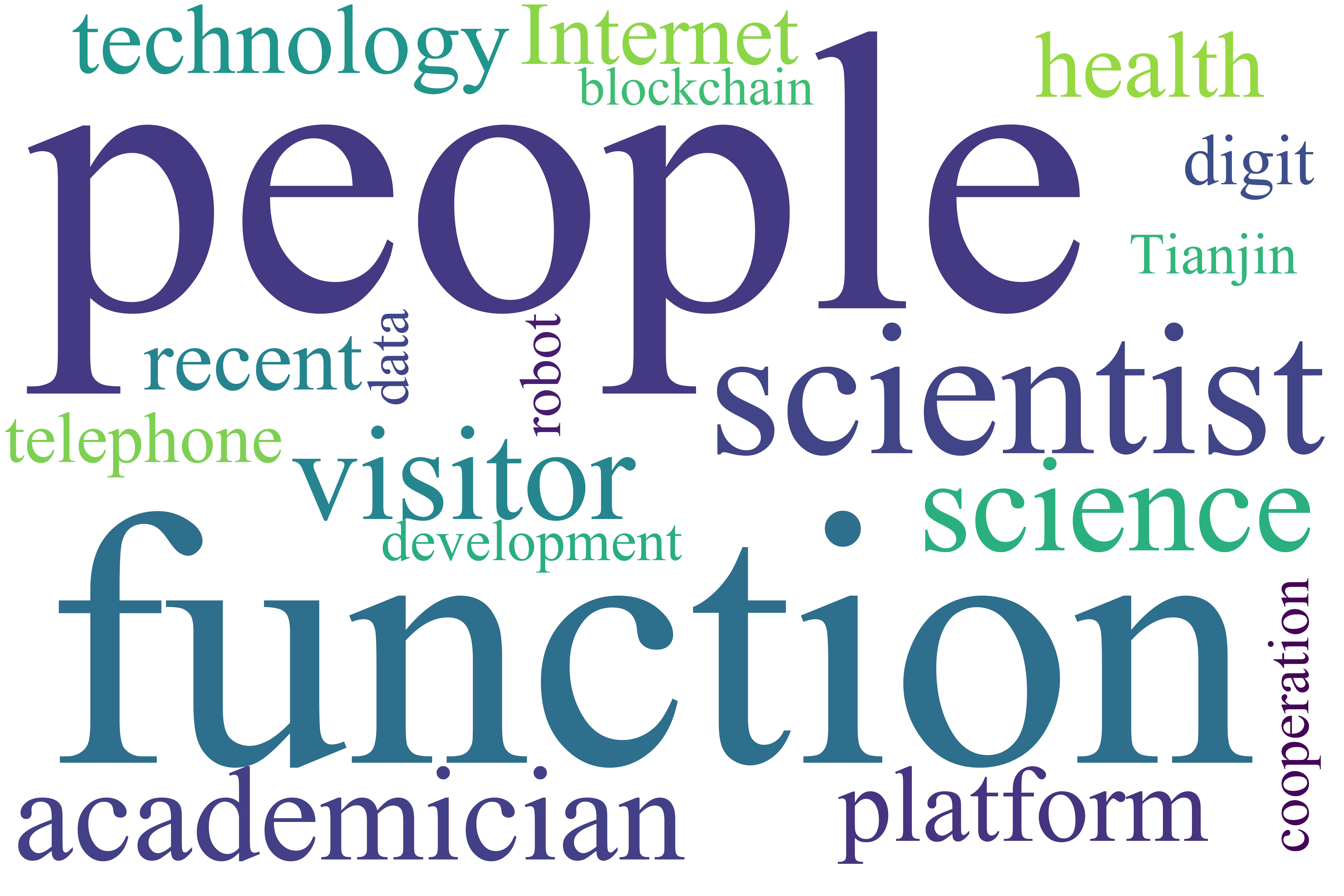}}
    \caption{Word clouds in four domains.}
    \label{fig:wordcloud}
\end{figure}

Weibo21 has multidimensional information related to news content, pictures, timestamp and comments. In order to quantify the differences among these domains, we perform some analyses. For example, news content is the most straightforward clue we can use for fake news detection. Therefore, we analysis the topic distribution of news among different domains. Figure \ref{fig:wordcloud} shows significant differences in the frequently used words. Due to limited space, we only demonstrate four of nine domains.
\vspace{-0.4 cm}
\section{MDFEND: Multi-domain Fake News Detection Model}
\begin{figure}
    \setlength{\belowcaptionskip}{-0.5cm}
    \setlength{\abovecaptionskip}{0cm}
    \centering
    \includegraphics[scale = 0.500]{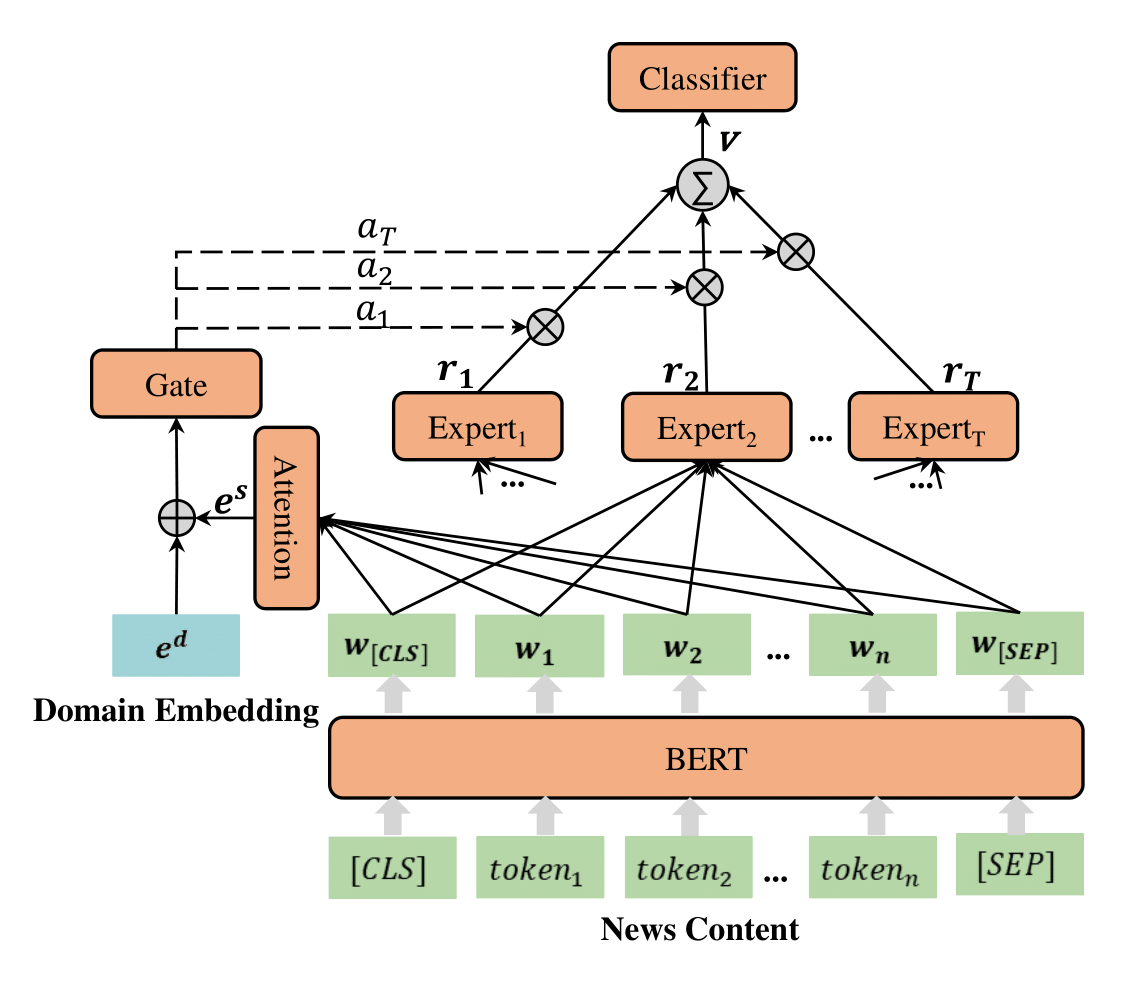}
    \caption{Overview of the proposed framework MDFEND.}
    \label{fig:model}
\end{figure}

In this section, we propose a novel framework, namely MDFEND, for MFND.
Same as the single domain methods, we treat multi-domain fake news detection as a binary classification problem. The overall framework is shown as Figure \ref{fig:model}.
\subsection{Representation Extraction}
For a piece of news, we tokenize its content with BertTokenizer~\cite{devlin2019bert}. After adding special tokens for classification ({i.e.,} $[CLS]$) as well as separation ({i.e.,} $[SEP]$), we obtain a list of tokens $[[CLS], token_1,  \newline... ,token_n, [SEP]]$ where $n$ is the number of tokens (words) in news content. These tokens are then fed into BERT to obtain word embeddings $\bm{W} = [\bm{w_{[CLS]}}, \bm{w_1}, ..., \bm{w_n}, \bm{w_{[SEP]}}]$, where all the word embeddings are processed by a Mask-Attention network to get the sentence-level embedding $\bm{e^s}$. To handle each domain specially, we define a learnable vector $\bm{e^d}$, namely domain embedding, to help  individualize representation extraction for each domain. Thus, a domain-specific value $\bm{e^d}$ will be learned for each domain.

With the advantage of Mixture-of-Expert~\cite{jacobs1991adaptive,ma2018modeling,zhu2021learning}, we employ multiple experts ({\it i.e.}, networks) to extract various representations of news. Intuitively, we can employ one expert to extract the news' representations for multiple domains. However, one expert only specializes in one area, therefore, the news' representation extracted by a single expert could only contain partial information, which cannot completely cover the characteristics of news content. As a result, we employ multiple experts for the sake of comprehensive. 

An ``expert" network can be denoted by $\Psi_i(\bm{W}; \theta_i)$  ($1 \leq i \leq T$), where $\bm{W}$ is a set of word embeddings as the input to the ``expert" network, $\theta_i$ represents the parameters to be learned, and  $T$ is a hyperparameter that indicates the number of expert networks. Let $r_i$ denote the output of an ``expert" network, {\it i.e.}, a representation extracted by the corresponding expert network. We have:
\begin{align}
    \bm{r_i} &= \Psi_i(\bm{W} ;\theta_i),
\end{align}
where each ``expert" network is a TextCNN~\cite{kim5882convolutionalneuralnetworksforsentence} in our design.
\vspace{-0.3cm}
\subsection{Domain Gate}
To obtain good performance on MFND, it is necessary to generate high-quality news representations that can represent news from different domains appropriately. Intuitively, we can average representations by all experts. However, the simple average operation will remove the domain-specific information, so the synthetic representation may not be good for MFND. Note that different experts specialize in different areas, and they are good at handling different domains. For MFND, we would like to select experts adaptively. 
Along this line, we propose a domain gate with the domain embedding as well as sentence embedding as input to guide the selection process. The output of the selection process is a vector $a$ indicating the weight ratio of each expert. We denote the domain gate as $G(\cdot ; \phi)$, and $\phi$ is the parameters in the domain gate:
\begin{align}
    \bm{a} &= softmax(G(\bm{e^d} \oplus \bm{e^s}; \phi)),
\end{align}
where the domain gate $G(\cdot ; \phi)$ is a feed-forward network. $\bm{e^d}$ and $\bm{e^s}$ are the domain embedding and sentence embedding, respectively. We use softmax function to normalize the output of $G(\cdot)$, and  $\bm{a} \in \mathbb{R}^n$ is the weight vector denoting the importance of different experts. With the domain gate, the news' final feature vector is obtained:
\begin{align}
    \bm{v} &= \sum_{i=1}^{T} a_i\bm{r_i}, 
\end{align}
\vspace{-0.6cm}
\subsection{Prediction}
The news' final feature vector is fed into the classifier, which is a multi-layer perception (MLP) network with a softmax output layer, for fake news detection:
\begin{align}
    \hat{y} &= softmax(MLP(\bm{v})), 
\end{align}
The goal of the fake news detector is to identify whether the news is fake or not. We use $y^i$ to represent actual value and $\hat{y}^i$ to represent predicted label. We employ Binary Cross-Entropy Loss (BCELoss) for classification:
\begin{align}
    L &= -\sum_{i=1}^N(y^i\log \hat{y}^i + (1-y^i)\log(1- \hat{y}^i)).
\end{align}
\begin{table*}[htbp]
  \centering
  \caption{Multi-domain Fake News Detection Performance (f1-score)}
  \scalebox{0.9}{
    \begin{tabular}{lcccccccccc}
    \toprule
    model & \multicolumn{1}{l}{Science} & \multicolumn{1}{l}{Military} & \multicolumn{1}{l}{Education} & \multicolumn{1}{l}{Accidents} & \multicolumn{1}{l}{Politics} & \multicolumn{1}{l}{Health} & \multicolumn{1}{l}{Finance} & \multicolumn{1}{l}{Entertainment} & \multicolumn{1}{l}{Society} & \multicolumn{1}{l}{All} \\
    \midrule
    
    TextCNN\_single &   0.7470    &   0.778    &   0.8882    &    0.8310   &   0.8694    &   0.9053    &   0.7909    &   0.8591    &   0.8727    &  0.8380 \\
    BiGRU\_single &   0.4876    &  0.7169     &   0.7067    &   0.7625    &   0.8477    &    0.8378   &   0.8109    &   0.8308    &   0.6067    &  0.7342\\
    BERT\_single &  0.8192     &   0.7795    &   0.8136    &   0.7885    &   0.8188    &   0.8909    &   0.8464    &   0.8638    &    0.8242   &  0.8272\\
    \midrule
    
    TextCNN\_all &   0.7254    &   0.8839    &   0.8362    &   0.8222    &  0.8561     &    0.8768   &   0.8638    &    0.8456   &    0.8540   &  0.8686\\
    BiGRU\_all  &   0.7269    &   0.8724    &   0.8138    &   0.7935    &   0.8356    &   0.8868    &    0.8291   &    0.8629   &   0.8485    & 0.8595 \\
    BERT\_all  &   0.7777    &   0.9072    &   0.8331    &   0.8512    &   0.8366    &    0.9090   &   0.8735    &    0.8769   &   0.8577    &  0.8795\\
    \midrule
    EANN  &   0.8225    &   0.9274    &   0.8624    &   0.8666    &  0.8705     &   0.9150    &   0.8710    &   0.8957    &   0.8877    &  0.8975\\
    MMOE  &   \textbf{0.8755}    &   0.9112    &   0.8706    &  0.8770     &   0.8620    &   0.9364    &   0.8567    &   0.8886    & 0.8750   &  0.8947 \\
    MOSE  &   0.8502    &   0.8858    &   0.8815    &   0.8672    &   0.8808    &   0.9179    &   0.8672    &   0.8913    &    0.8729   &  0.8939\\
    EDDFN &   0.8186    &    0.9137   &   0.8676    &   0.8786    &    0.8478   &   0.9379    &   0.8636    &   0.8832    &   0.8689    &  0.8919\\
    \midrule
    \textbf{MDFEND} &   0.8301    &   \textbf{0.9389}    &   \textbf{0.8917}    &   \textbf{0.9003}    & \textbf{0.8865}   &   \textbf{0.9400}    &   \textbf{0.8951}    &   \textbf{0.9066}    &    \textbf{0.8980}   &  \textbf{0.9137} \\
    \bottomrule
    \end{tabular}%
  }
  \label{tab:results}%
\end{table*}%

\vspace{-0.4cm}
\section{Experiment}
In this section, we evaluate the effectiveness of our proposed MDFEND framework, and compare with other baselines on our dataset.
\vspace{-0.6 cm}
\subsection{Baseline Methods}

There are three types of baselines in our experiments: (1) \textbf{single-domain baselines}: TextCNN\_single~\cite{kim5882convolutionalneuralnetworksforsentence}, BiGRU\_single~\cite{ma2016detecting}, and BERT\_single~\cite{devlin2019bert}; (2) \textbf{mixed-domain baselines}: TextCNN\_all~\cite{kim5882convolutionalneuralnetworksforsentence}, BiGRU\_all~\cite{ma2016detecting} and BERT\_all~\cite{devlin2019bert}; (3) \textbf{multi-domain baselines}: EANN~\cite{wang2018eann}, MMOE~\cite{ma2018modeling}, MOSE~\cite{qin2020multitask} and EDDFN~\cite{silva2021embracing}. In \textbf{single-domain baselines}, we conduct experiment with one model on one single domain at a time (e.g., train \textbf{TextCNN\_single} on \textbf{Science} domain), and the results in the last column are the average of the ones in the former columns. In \textbf{mixed-domain baselines}, we perform experiment with one model on all domains at a time (e.g., train \textbf{TextCNN\_all} on \textbf{all} domains), and calculate the f1-score of each domain respectively, while the results in the last column are not just the average of the ones in the former columns as in single domain baselines, but calculated using data from all domains. Models used in \textbf{multi-domain baselines} combine data from different domains according to their structures.

To make the baseline models work for MFND, based on their original designs, we made  the following modifications in our experiments.
\textbf{BERT}~\cite{devlin2019bert}: In our experiments, we freeze all layers in BERT, and average the word embeddings in the last layer as the sentence representation. A multi-layer perception (MLP) is stacked on the top to perform binary classification. \textbf{TextCNN}~\cite{kim5882convolutionalneuralnetworksforsentence}: We employ the same convolutional structure as in our expert module. The input feature of TextCNN is embedded by word2vec. \textbf{BiGRU}~\cite{ma2016detecting}: Different from ~\cite{ma2016detecting}, we model each news piece as a sequential input to the BiGRU in order to preserve the sequential information. 
\textbf{EANN}~\cite{wang2018eann}: We use this model to extract domain-independent features. We only use the textual branch, which is consistent with TextCNN in model structure. \textbf{MMOE}~\cite{ma2018modeling} and \textbf{MOSE}~\cite{qin2020multitask}: These two models are proposed for multi-task learning. We assume that fake news detection in multiple domains are different tasks, and use the two models for multi-domain fake news detection. \textbf{EDDFN}~\cite{silva2021embracing}: it is a model for cross-domain fake news detection, which models different domains implicitly and jointly preserves domain-specific and cross-domain knowledge. In our experiments, we abandon the domain discovery module and use our manually labeled domains for multi-domain fake news detection.
\vspace{-0.25 cm}
\subsection{Experiment Setting}
The details of our dataset Weibo21 are listed in Table \ref{tab:data_statistic}.
For a fair comparison, we set the same parameters for all methods. The MLP in these models uses the same structure with one dense layer (384 hidden units). For all experiments, the max length of the sentence is 170, and the dimension of embedding vectors of words is fixed to 768 for BERT~\cite{devlin2019bert} and 200 for Word2Vec~\cite{le2014distributed,mikolov2013efficient}. We employ the Adam~\cite{kingma2015adam} optimizer and search its learning rate from 1e-6 to 1e-2. For all methods, the mini-batch size is 64. In order to enhance the credibility of our experiments, the process is performed for 10 times and the average f1-score is reported.
\vspace{-0.2 cm}
\subsection{Results}
We compare our MDFEND model with single-domain, mixed-domain and multi-domain methods to testify the effectiveness. The results are shown in Table \ref{tab:results}. Experimental results further reveal several insightful observations.

(1) Mixed-domain models and multi-domain models work much better than single-domain models in general, which demonstrates that additional data is of great importance. 

(2) Compared with mixed-domain models, multi-domain models perform better in general. This illustrates that multi-domain learning is useful and necessary for MFND.

(3) Note that there are single-domain models performing better on a specific domain than the corresponding mixed-domain model (e.g., TextCNN\_single v.s. TextCNN\_all on Health domain), which illustrates that simply combining data from different domains may result in negative effect from additional data.

(4) MFND models domain relationships better by feeding both domain embedding and content to the gate. Moreover, combining multiple domains softly is better than decouple domain-shared and domain-specific features roughly (EDDFN). Therefore, MDFEND model performs better than other multi-domain models.
\vspace{-0.1 cm}
\section{Conclusion}
In this paper, we study the problem of multi-domain fake news detection (MFND). 
We construct Weibo21, a MFND dataset. To the best of our knowledge, it is the first MFND dataset collected from one platform with richest domains; We propose a simple but effective method named MDFEND for MFND, which utilizes domain gate to aggregate multiple representations extracted by mixture-of-experts; We also systematically evaluated MFND performance with different methods on the proposed Weibo21 dataset, and experiments show the effectiveness of our MDFEND model.
\begin{acks}
This work was supported by the Zhejiang Provincial Key Research and Development Program of China (2021C01164), and the National Natural Science Foundation of China (U1703261).
\end{acks}

\balance
\bibliographystyle{ACM-Reference-Format}
\bibliography{sample-base}

\end{document}